# An End-to-End Multi-Task Learning Model for Image-based Table Recognition

Nam Tuan Ly[a], and Atsuhiro Takasu[b]
*National Institute of Informatics (NII), Tokyo, Japan*
*{namly, takasu}@nii.ac.jp*

Abstract: Image-based table recognition is a challenging task due to the diversity of table styles and the complexity of table structures. Most of the previous methods focus on a non-end-to-end approach which divides the problem into two separate sub-problems: table structure recognition; and cell-content recognition and then attempts to solve each sub-problem independently using two separate systems. In this paper, we propose an end-to-end multi-task learning model for image-based table recognition. The proposed model consists of one shared encoder, one shared decoder, and three separate decoders which are used for learning three sub-tasks of table recognition: table structure recognition, cell detection, and cell-content recognition. The whole system can be easily trained and inferred in an end-to-end approach. In the experiments, we evaluate the performance of the proposed model on two large-scale datasets: FinTabNet and PubTabNet. The experiment results show that the proposed model outperforms the state-of-the-art methods in all benchmark datasets. The code is available at https://github.com/namtuanly/MTL-TabNet.

## 1 INTRODUCTION

The tabular format is one of the rich-information formats and is widely used in communication, research, and data analysis. Tables commonly appear in research papers, books, handwritten notes, invoices, financial documents, and many other places. Thus, table understanding becomes one of the essential techniques in document analysis systems and attracts the attention of numerous researchers.

Image-based table recognition is the key step of table understanding which refers to the representation of a table image in a machine-readable format, where its structure and the content within each cell are encoded according to a pre-defined standard (Zhong et al., 2020). The machine-readable format can be HTML code (Jimeno Yepes et al., 2021; Li et al., 2019; Zhong et al., 2020) or LaTeX code (Deng et al., 2019; Kayal et al., 2021). The choice of the machine-readable format is ultimately not very important, since one can be transformed into the other. Image-based table recognition is a challenging task due to the diversity of table styles and the complexity of table structures. In the past few decades, many table recognition methods have been proposed and can be divided into two categories: end-to-end methods and non-end-to-end methods. Most of the previous works (Nassar et al., 2022; Qiao et al., 2021; Ye et al., 2021) focus on non-end-to-end approaches which divide the problem into two separate sub-problems: table structure recognition; and cell-content recognition, and then attempt to solve each sub-problem independently using two separate systems. On the other hand, the end-to-end approach attempts to solve the problem using a single model (specifically a Deep Neural Network) and achieves state-of-the-art results on many tasks such as machine translation (Vaswani et al., 2017), speech recognition (Bahdanau et al., 2016), and text recognition (Lu et al., 2021; Ly et al., 2021). However, to the best of our knowledge, there are few studies (Deng et al., 2019; Zhong et al., 2020) on the end-to-end approach to table recognition and their performance is mediocre compared to the non-end-to-end methods.

In this paper, we formulate the problem of table recognition as a multi-task learning problem, which requires the model to be jointly learned in three sub-tasks of table recognition. To address this problem, we propose a novel end-to-end multi-task learning model which consists of one shared encoder, one shared decoder, and three separate decoders for three sub-tasks of table recognition: table structure recognition, cell detection, and cell-content recognition. The model takes an input table image and produces the table structure information, location of table cells, and contents of table cells, which can be easily transformed into the HTML code (or LaTeX code) representing the table. The shared components are repeatedly trained from the gradients received from three sub-tasks while each of three separate decoders is trained from the gradients of its task. The whole system can be easily trained and inferred in an end-to-end approach.

---

[a] https://orcid.org/0000-0002-0856-3196
[b] https://orcid.org/0000-0002-9061-7949

We have evaluated the performance of our model on the PubTabNet (Zhong et al., 2020) and FinTabNet (Zheng et al., 2021) datasets, demonstrating that our model outperforms state-of-the-art methods in both table structure recognition and table recognition. We further evaluated our model on the final evaluation set of Task-B in the ICDAR 2021 competition on scientific literature parsing (Jimeno Yepes et al., 2021) (ICDAR 2021 competition in short), demonstrating that our model achieves competitive results when compared to the top three solutions. The code will be publicly released to GitHub.

In summary, the main contributions of this paper are as follows:
- We present a novel end-to-end multi-task learning model for image-based table recognition. The proposed model can be easily trained and inferred in an end-to-end approach.
- Across all benchmark datasets, the proposed model outperforms the state-of-the-art methods.
- Although we used neither any additional training data nor ensemble techniques, our model achieves competitive results when compared to top three solutions in ICDAR2021 competition.

The rest of this paper is organized as follows. In Sec.2, we give a brief overview of the related works. In Sec. 3, we introduce the overview of the proposed model. In Sec. 4, we report the experimental details and results. Finally, we draw conclusions in Sec. 5.

## 2 RELATED WORK

Table understanding in unstructured documents can be defined in three steps: 1) table detection: detecting the bounding boxes of tables in documents (Casado-García et al., 2020; Huang et al., 2019); 2) table structure recognition: recognizing the structural information of tables (Itonori, 1993; Kieninger, 1998; Wang et al., 2004); 3) table recognition: recognizing both the structural information and the content within each cell of tables (Deng et al., 2019; Ye et al., 2021; Zhong et al., 2020). We will briefly survey table structure recognition, and then table recognition.

**Table structure recognition** can be considered the first step of table recognition and has been studied by researchers around the world for the past few decades. Early works of table structure recognition are based on hand-crafted features and heuristic rules (Itonori, 1993; Kieninger, 1998; Wang et al., 2004). These methods are mostly applied to a simple structure or pre-defined data formats. In recent years, inspired by the success of deep learning in various tasks, especially object detection and semantic segmentation, many deep learning-based methods (Raja et al., 2020; Schreiber et al., 2017) have been presented to recognize table structures. S. Schreiber et al. (Schreiber et al., 2017) proposed a two-fold system named DeepDeSRT that applies Faster RCNN (Ren et al., 2015) and FCN(Long et al., 2015) for both table detection and row/column segmentation. Sachin et al. (Raja et al., 2020) presented a table structure recognizer named TabStruct-Net that combines cell detection and interaction modules to localize the cells and predicts their row and column associations with other detected cells. Structural constraints are incorporated as additional differential components to the loss function for cell detection. Recently, graph neural networks are also used for table structure recognition by encoding document images as graphs (Qasim et al., 2019).

**Table recognition:** Most of the previous works of table recognition (Nassar et al., 2022; Qiao et al., 2021; Ye et al., 2021; Zhang et al., 2022) focus on non-end-to-end approaches which divide the problem into two separate sub-problems: table structure recognition; and cell-content recognition, and then attempt to solve each sub-problem independently using two separate systems. J. Ye et al. (Ye et al., 2021) proposed a Transformer-based model named TableMASTER for table structure recognition and combined it with a text line detector to detect text lines in each table cell. Finally, they employed a text line recognizer based on (Lu et al., 2021) to recognize each text line. Their system achieved second place in ICDAR2021 competition. A. Nassar et al. (Nassar et al., 2022) proposed a Transformer-based model named TableFormer for recognizing both table structure and the bounding box of each table cell and then using these bounding boxes to extract the cell contents from the PDF to build the whole table recognition system. Z. Zhang et al. (Zhang et al.,

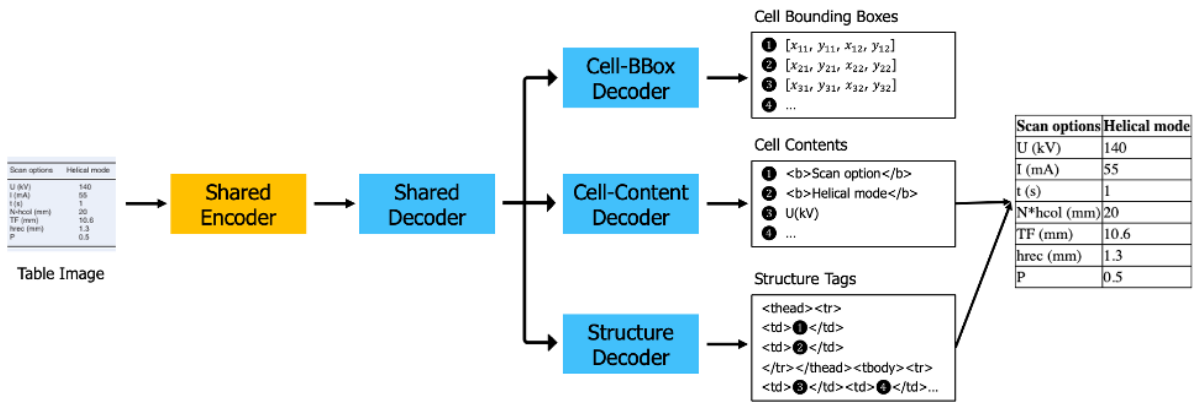

Figure 1. The overview of the proposed model.

2022) proposed a table structure recognizer, Split, Embed, and Merge (SEM) for recognizing the table structure. Then, they combined SEM with an attention-based text recognizer to build the table recognizer and achieved third place in the ICDAR2021 competition.

Recently, due to the rapid development of deep learning and the increase in the tabular data, some works (Deng et al., 2019; Zhong et al., 2020) try to focus on end-to-end approaches. However, their performance is still mediocre compared to the non-end-to-end methods. Y. Deng et al. (Deng et al., 2019) formulated table recognition as the image to latex problem and employed IM2TEX (Deng et al., 2016) model for table recognition. X. Zhong et al (Zhong et al., 2020) proposed an encoder-dual-decoder (EDD) model for recognizing both table structure and content of each cell. They also publicized a table recognition dataset PubTabNet to the community.

In 2021, IBM Research in conjunction with IEEE ICDAR held ICDAR2021 competition on scientific literature parsing (Jimeno Yepes et al., 2021) (ICDAR2021 competition in short). The competition consists of two tasks: Task A - Document layout recognition which identifies the position and category of document layout elements, including title, text, figure, table, and list; and Task B - Table recognition which converts table images into HTML code. For Task B, there are 30 submissions from 30 teams for the Final Evaluation Phase and most of the top 10 systems are non-end-to-end approaches and employ ensemble techniques to improve their performance.

## 3 METHODOLOGY

The proposed model consists of one shared encoder, one shared decoder, and three separate decoders for three sub-tasks of the table recognition problem as shown in Fig. 1. The shared encoder encodes the input table image as a sequence of features. The sequence of features is passed to the shared decoder and then the structure decoder to predict a sequence of HTML tags that represent the structure of the table. When the structure decoder produces the HTML tag representing a new cell ('<td>' or '<td …'), the output of the shared decoder corresponding to that cell and the output of the shared encoder are passed into the cell-bbox decoder and the cell-content decoder to predict the bounding box coordinates and the text content of that cell. Finally, the text contents of cells are inserted into the HTML structure tags corresponding to their cells to produce the final HTML code of the input table image. Fig. 2 shows the detail of the five components in our model. We describe the detail of each component in the following sections.

### 3.1 Shared Encoder

In this work, we use a CNN backbone network as the feature extractor followed by a positional encoding layer to build the shared encoder. The feature extractor extracts visual features from an input table image of the size $\{h, w, c\}$ ($c$ is the color channel), resulting in a feature grid $F$ of the size $\{h', w', k\}$ ($k$ is the number of the feature maps). Then the feature grid $F$ is unfolded into a sequence of features (column by column from left to right in each feature map) before being fed into the positional encoding layer to get the encoded sequence of features. The encoded sequence of features will be fed into the shared decoder and three separate decoders.

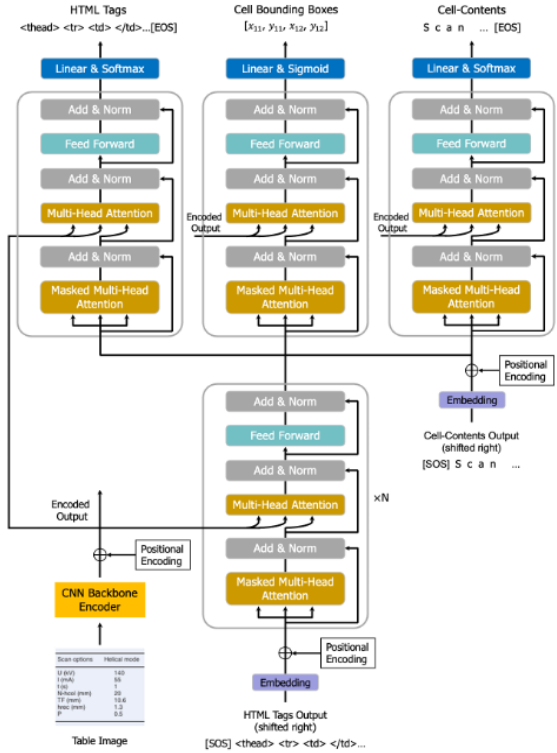

Figure 2. Network architecture of the proposed model.

## 3.2 Shared Decoder

The architecture of all decoders in our model is inspired by the original Transformer decoder (Vaswani et al., 2017) which is composed of a stack of *N* identical Transformer decoder layers where *N* can be a hyperparameter. Each identical Transformer decoder layer (identical layer in short) has three sub-layers: a multi-head self-attention mechanism; a masked multi-head self-attention mechanism; and a position-wise fully connected feed-forward network, and helps the decoders focus on appropriate places in the input table image.

At the top of the shared encoder, the shared decoder is composed of a stack of *N=2* identical layers. As shown in Fig.2, the output of the shared encoder is fed into the multi-head self-attention mechanism of each identical layer as the value and key vectors. During training, the right-shifted sequence of target HTML tags (structural tokens) of the table structure (after passing through the embedded layer and the positional encoding layer) is passed into the bottom of the shared decoder as the query vector. In the inference stage, the right-shifted sequence of target HTML tags is replaced by the right-shifted sequence of HTML tags outputted by the structure decoder. Finally, the outputs of the shared decoder will be fed into the three separate decoders to predict three sub-tasks of the table recognition problem.

## 3.3 Structure Decoder

At the top of the shared decoder, the structure decoder uses the outputs of the shared decoder and the outputs of the shared encoder to predict a sequence of HTML tags of the table structure. Inspired by the works in (Ye et al., 2021; Zhong et al., 2020), the HTML tags of the table structure are tokenized at the HTML tag level except for the tag of a cell. The form of '<td></td>' is treated as one token class and the tag of the spanning cells is broken down into '<td', 'rowspan=' or 'colspan=', with the number of spanning cells, and '>'. Thus, the structural token of '<td></td>' or '<td' represents a new table cell. As shown in Fig. 2, the structure decoder is composed of one identical layer followed by a linear layer and a softmax layer. The identical layer takes the outputs of the shared decoder as the query vector input and the outputs of the shared encoder as the key and value vector inputs. The output of the identical layer is fed into the linear layer, and then the softmax layer to generate the sequence of structural tokens.

## 3.4 Cell-BBox Decoder

When the structure decoder generates a structural token representing a new cell ('<td></td>' or '<td'), the cell-bbox decoder is triggered and uses the output of the shared decoder corresponding to this cell to predict the bounding box coordinates of this cell.

As shown in Fig. 2, we use one identical layer followed by a linear layer and a sigmoid layer to build the cell-bbox decoder. The identical layer takes the output of the shared decoder and the output of the shared encoder as the input and learns to focus on appropriate places in the input image. The output of the identical layer is fed into the linear layer and then the sigmoid layer to predict the four coordinates of the cell bounding box.

## 3.5 Cell-Content Decoder

Similar to the cell-bbox decoder, the cell-content decoder selects the outputs of the shared decoder referring to the structural tokens representing a new cell ('<td></td>' or '<td') and uses them to recognize the text contents of cells. The cell-content decoder in the proposed model can be considered a text recognizer and the text output are tokenized at the character level. In this work, we use one identical layer followed by a linear layer and a softmax layer to build the cell-content decoder as shown in Fig. 2. The output of the shared encoder are fed into the identical layer as the input value and key vectors. During training, the right-shifted target of the cell content (the right-shifted output of the cell-content decoder in the testing phase) is passed through the embedded and the positional encoding layers and then added to the output of the shared decoder before being fed into the identical layer as the query vector. Finally, the output of the identical layer is fed into the linear layer and then the softmax layer to generate the cell content.

### 3.6 Network Training

In our model, the shared components are repeatedly trained from the gradients received from three subtasks while each of three separate decoders is trained from the gradients obtained from its task. The whole system can be trained end-to-end on pairs of table images and their annotations of the table structure, the text content, and its bounding box per non-empty table cell by stochastic gradient descent algorithms. The overall loss of our model is defined as the following:

$$\mathcal{L} = \lambda_1 \mathcal{L}_{\text{struc.}} + \lambda_2 \mathcal{L}_{\text{cont.}} + \lambda_3 \mathcal{L}_{\text{bbox}} \quad (1)$$

where $\mathcal{L}_{\text{struc.}}$ and $\mathcal{L}_{\text{cont.}}$ are the table structure recognition loss and the cell-content prediction loss, respectively that are implemented in Cross-Entropy loss, $\mathcal{L}_{\text{bbox}}$ is the cell-bbox regression loss which is optimized by L1 loss. $\lambda_1$, $\lambda_2$, and $\lambda_3$ are weight hyperparameters.

## 4 EXPERIMENTS

To evaluate the performance of the proposed model, we conducted experiments on two datasets: FinTabNet (Zheng et al., 2021) and PubTabNet (Zhong et al., 2020). The information of the datasets is given in Sec 4.1. The implementation details are described in Sect. 4.2; the experimental results are presented in Sect. 4.3; and the visualization results are shown in Sect. 4.4.

### 4.1 Datasets

In this paper, we conduct the experiments on the following two large-scale datasets that contain the annotations of both the structure of the table and the text content with the position of each non-empty table cell.

**PubTabNet** (Zhong et al., 2020) is a large-scale table image dataset that contains over 568k samples with their corresponding annotations of the table structure presented in HTML format, the text content, and its bounding box per non-empty table cell. This dataset is created by collecting scientific articles from PubMed Central Open Access Subset (PMCOA). The dataset is used in the ICDAR2021 competition (Jimeno Yepes et al., 2021) and divided into 500,777 training samples and 9,115 validation samples in the development phase, and 9,064 final evaluation samples in the Final Evaluation Phase.

**FinTabNet** is another large-scale table image dataset published by X. Zheng et al. (Zheng et al., 2021). The dataset is composed of complex tables from the annual reports of the S&P 500 companies with detailed annotations of the table structure and table cell information like the PubTabNet dataset. This dataset consists of 112k table images which are divided into training, testing, and validation sets with a ratio of 81% : 9.5% : 9.5%.

### 4.2 Implementation Details

We use the ResNet-31 network (He et al., 2016) to build the CNN backbone in our model. To enable the CNN backbone network to model the global context from the input image, we add the Multi-Aspect Global Context Attention (GCAttention) proposed by Ning Lu et. al. (Lu et al., 2021) after each residual block of the ResNet-31 network. All images are resized to 480*480 pixels and the feature map outputted from the CNN backbone has a dimension of 60*60.

At the decoders, all identical layers have the same architecture with the input feature size of 512, the feed-forward network size of 2048, and 8 attention heads. The maximum length of a sequence of structural tokens in the structure decoder is 500 and the maximum length of a sequence of cell tokens in the cell-content decoder is 150. We empirically set all weight hyperparameters as $\lambda_1 = \lambda_2 = \lambda_3 = 1$.

Our model is implemented with Pytorch and the MMCV library (MMCV Contributors, 2018). The model is trained on two NVIDIA A100 80G with a batch size of 4 in each GPU. The initializing learning rate is 0.001 for the first 12 epochs. Afterward, we reduce the learning to 0.0001 and train for 8 more epochs or convergence.

## 4.3 Experimental Results

To evaluate the performance of the proposed model, we employ the Tree-Edit-Distance-Based Similarity (TEDS) metric as defined in (Zhong et al., 2020). We also denote TEDS-struc. as the TEDS score between two tables when considering only the table structure information.

### 4.3.1 Table Structure Recognition

First, we conducted experiments to verify the effectiveness of the proposed model for recognizing the table structure. Table 1 compares the table structure recognition performance (TEDS-struc. scores) of the proposed model and the previous table structure recognition methods on PubTabNet and FinTabNet datasets.

As shown in Table 1, our model achieves superior performance on all benchmark datasets compared to the state-of-the-art models. Specifically, with TEDS-struc. of 98.79% on the FinTabNet dataset, our model improves TableFormer (Nassar et al., 2022) by 2% and other methods by more than 7.7%, and even GTE (FT) (Zheng et al., 2021) is pretrained in the PubTabNet dataset. On the PubTabNet dataset, our model achieved TEDS-struc. of 97.88% which again improves TableFormer and LGPMA (Qiao et al., 2021) by about 1.1% and other methods by more than 4.87%. Note that LGPMA requires additional annotation information for training. The proposed model also achieves state-of-the-art accuracies on complex tables (tables with multi-column or multi-row cells) of both datasets.

**Cell Detection:** Like any object detection model, the cell-bbox decoder produces the bounding boxes of the cells which can be evaluated by the PASCAL VOC mAP metric. Table 2 shows the mAP of the proposed model in comparison with the previous works in (Nassar et al., 2022) on the PubTabNet dataset. Our model achieves the state-of-the-art results and significantly improves TableFormer and EDD + BBox by more than 6.8%. Even without post-processing, the proposed model slightly outperforms TableFormer + PP which uses the information of the cell bounding boxes from the PDF document in post-processing.

Table 1: Table structure recognition results on PubTabNet validation set (PTN) and FinTabNet (FTN).

| Dataset | Model | TEDS-struc. (%) | | |
|---|---|---|---|---|
| | | Sim. | Com. | All |
| FTN | EDD (Zhong et al., 2020) | 88.40 | 92.08 | 90.60 |
| | GTE (Zheng et al., 2021) | - | - | 87.14 |
| | GTE (FT) (Zheng et al., 2021) | - | - | 91.02 |
| | TableFormer (Nassar et al., 2022) | 97.50 | 96.00 | 96.80 |
| | **Our Model** | **99.07** | **98.46** | **98.79** |
| PTN | EDD (Zhong et al., 2020) | 91.10 | 88.70 | 89.90 |
| | GTE (Zheng et al., 2021) | - | - | 93.01 |
| | LGPMA (Qiao et al., 2021) | - | - | 96.70 |
| | TableFormer (Nassar et al., 2022) | 98.50 | 95.00 | 96.75 |
| | **Our Model** | **99.05** | **96.66** | **97.88** |

Sim. (Simple): Tables without multi-column or multi-row cells.
Com. (Complex): Tables with multi-column or multi-row cells.
(FT) Model was trained on PubTabNet and then finetuned.

Table 2: Cell detection results on PubTabNet validation set. PP: Post-processing.

| Model | mAP (%) |
|---|---|
| EDD + BBox | 79.20 |
| TableFormer | 82.10 |
| EDD + BBox + PP | 82.70 |
| TableFormer + PP | 86.80 |
| **Our Model** | **88.93** |

### 4.3.2 Results of Table Recognition

In this experiment, we evaluate the performance of the proposed model in the table recognition problem which recognizes both the structure of the table and the content within each cell. Table 3 shows table recognition results of the proposed model in comparison with the previous table recognition methods on the PubTabNet dataset. Our model outperforms all the previous methods without ensemble techniques. Specifically, our model achieved TEDS of 96.67% which improves VCGoup's solution (Ye et al., 2021) by 0.41%, LGPMA + OCR (Qiao et al., 2021) by 2%, and others by more than 3%. Note that VCGoup's solution, and SEM (Zhang et al., 2022) are the 2nd ranking, and 3rd ranking solutions in ICDAR2021 competition. LGPMA (Qiao et al., 2021) is the table structure recognizer component in the 1st ranking solution in

Table 3: Table recognition results on PubTabNet validation set.

| Model | TEDS (%) | | |
|---|---|---|---|
| | Sim. | Com. | All |
| EDD (Zhong et al., 2020) | 91.20 | 85.40 | 88.30 |
| TabStruct-Net (Raja et al., 2020) | - | - | 90.10 |
| GTE (Zheng et al., 2021) | - | - | 93.00 |
| TableFormer (Nassar et al., 2022) | 95.40 | 90.10 | 93.60 |
| SEM [3] (Zhang et al., 2022) | 94.80 | 92.50 | 93.70 |
| LGPMA + OCR [1] (Qiao et al., 2021) | - | - | 94.60 |
| VCGoup [2] (Ye et al., 2021) | - | - | 96.26 |
| **Our Model** | **97.92** | **95.36** | **96.67** |
| VCGoup + ME [2] (Ye et al., 2021) | - | - | 96.84 |

(1)(2)(3) are 1st, 2nd, and 3rd ranking solutions in ICDAR2021 competition. ME: Model Ensemble.

Table 4: Table recognition results on PubTabNet final evaluation set.

| Team Name | TEDS (%) | | |
|---|---|---|---|
| | Simp. | Comp. | All |
| Davar-Lab-OCR | 97.88 | 94.78 | **96.36** |
| VCGroup | **97.90** | 94.68 | 96.32 |
| XM | 97.60 | **94.89** | 96.27 |
| **Our Model** | 97.60 | 94.68 | 96.17 |
| YG | 97.38 | 94.79 | 96.11 |
| DBJ | 97.39 | 93.87 | 95.66 |
| TAL | 97.30 | 93.93 | 95.65 |
| PaodingAI | 97.35 | 93.79 | 95.61 |
| anyone | 96.95 | 93.43 | 95.23 |
| LTIAYN | 97.18 | 92.40 | 94.84 |

ICDAR2021 competition. All other methods except EDD (Zhong et al., 2020) are non-end-to-end approach and the methods in (Qiao et al., 2021; Ye et al., 2021; Zhang et al., 2022) requires additional annotation information for training.

Without ensemble technique as well as additional annotation information, however, our model achieves the competitive results when compared to VCGoup's solution + ME in (Ye et al., 2021) which requires annotations of text-line bounding boxes of cell contents in table images and employs three model ensembles in the table structure recognition and three model ensembles in the text line recognition.

We also evaluate our model on the final evaluation set of the PubTabNet dataset which is used for the Final Evaluation Phase in ICDAR2021 competition. Table 4 compares TEDS scores by our model and the top 10 solutions in ICDAR2021 competition (Jimeno Yepes et al., 2021). Although we used neither any additional training data nor ensemble techniques, our model outperforms the 4th ranking solution named YG and achieves competitive results when compared to the top three solutions in the final evaluation set of Task-B in the ICDAR 2021 competition. Furthermore, the proposed model achieves a similar TEDS score on complex tables with the 2nd ranking solution named VCGroup. Note that the 1st ranking solution is a non-end-to-end approach which employs LGPMA (Qiao et al., 2021) to recognize the structure of the table and then uses attention-based text recognizer to provide the OCR information of the table cells. They also adopt multi-scale ensemble strategy to further improve the performance. The other 9 solutions also are non-end-to-end approaches and most of them use additional data for training as well as ensemble methods.

### 4.4 Visualization Results

In this section, we show some visualization results of the proposed model on the PubTabNet dataset. As shown in Fig. 3, the left image is the original image with detected bounding boxes of table cells, and the right image is the predicted HTML code of the table view on the web browser. As it is shown, our model is able to predict complex table structure as well as bounding boxes and contents for all table cells, even for the empty cells or cells that cross span multiple rows/columns.

## 5 CONCLUSIONS

In this paper, we formulate the problem of table recognition as a multi-task learning problem and propose a novel end-to-end multi-task learning model which consists of one shared encoder, one shared decoder, and three separate decoders for three sub-tasks of table recognition: table structure recognition, cell detection, and cell-content recognition. The shared components are repeatedly trained from the gradients received from three sub-tasks while each of three separate decoders is trained from the gradients of its task. Extensive experiments on two large-scale datasets demonstrate our model achieved state-of-the-art accuracies in both table structure recognition and table recognition. Although we used neither any additional training data nor ensemble techniques, our model outperforms the 4th ranking solution and achieves competitive results when compared to the top three solutions in ICDAR 2021 competition.

Figure 3. Visualization results on PubTabNet. Green boxes are detected bounding boxes of table cells, and the right image is the predicted HTML code of the table viewed on the web browser.

In the future, we will conduct experiments of the proposed model on the table image datasets of other languages. We also plan to incorporate language models into the structure decoder as well as the cell-content decoder to improve the performance of the proposed model.

## ACKNOWLEDGEMENTS

This work was supported by the Cross-ministerial Strategic Innovation Promotion Program (SIP) Second Phase and "Big-data and AI-enabled Cyberspace Technologies" by New Energy and

Industrial Technology Development Organization (NEDO).